\documentclass[conference]{IEEEtran}
\IEEEoverridecommandlockouts
\usepackage{url}
\usepackage{amsmath, amssymb, amsthm}
\usepackage{wrapfig}
\newcommand{\comment}[1]{}

\usepackage[T1]{fontenc}    
\usepackage{hyperref}  
\usepackage{cleveref}
\crefname{equation}{}{}
\Crefname{equation}{}{}
\usepackage{url}            
\usepackage{booktabs}       
\usepackage{amsfonts}       
\usepackage{nicefrac}       
\usepackage{microtype}      
\usepackage{algorithm}
\usepackage[noend]{algpseudocode}
\usepackage{amssymb,amsmath, amsthm}
\usepackage[utf8]{inputenc}
\usepackage{mathtools}
\usepackage{subcaption}
\usepackage{graphicx}
\usepackage{caption}
\usepackage{bbm}
\usepackage{wrapfig}
\usepackage[shortlabels]{enumitem}
\usepackage{array}
\usepackage{arydshln}

\usepackage[colorinlistoftodos,prependcaption]{todonotes}

\definecolor{gr}{rgb}{0.25, 0.25, 0.25}

\newcommand{\ec}{K}
\newcommand{\scc}{m}
\newcommand{\ic}{k}

\newcommand{\ucbc}{\textsc{UCB-CS}}

\newcommand{\gf}{F}

\newcommand{\lff}{f}
\newcommand{\lf}{F_\ic}

\newcommand{\wb}{\mathbf{w}}

\newcommand{\bdat}{\mathcal{B}_\ic}

\newcommand{\ldat}{D_\ic}

\newcommand{\rdat}{p_k}

\newcommand{\ti}{(t)}
\newcommand{\tip}{(t+1)}

\newcommand{\midat}{\xi_{\ic}^{\ti}}

\newcommand{\sgrad}{g_\ic}

\newcommand{\lr}{\eta_t}

\newcommand{\owglb}{\overline{\wb}}

\newcommand{\pir}{\pi_{\text{rand}}}

\newcommand{\piucb}{\pi_{\text{ucb-cs}}}

\newcommand{\pipd}{\pi_{\text{pow-d}}}

\newcommand{\piprr}{\pi_{\text{rpow-d}}}

\DeclarePairedDelimiter\floor{\lfloor}{\rfloor}

\newcommand{\st}{\mathcal{S}^{\ti}}

\renewcommand{\mod}{~\mathrm{mod}~}

\algnewcommand\algorithmicswitch{\textbf{switch}}
\algnewcommand\algorithmiccase{\textbf{case}}
\let\oldReturn\Return
\renewcommand{\Return}{\State\oldReturn}
\algdef{SE}[SWITCH]{Switch}{EndSwitch}[1]{\algorithmicswitch\ #1\ \algorithmicdo}{\algorithmicend\ \algorithmicswitch}%
\algdef{SE}[CASE]{Case}{EndCase}[1]{\algorithmiccase\ #1}{\algorithmicend\ \algorithmiccase}%
\algtext*{EndSwitch}%
\algtext*{EndCase}%

\def\BibTeX{{\rm B\kern-.05em{\sc i\kern-.025em b}\kern-.08em
    T\kern-.1667em\lower.7ex\hbox{E}\kern-.125emX}}

\begin{document}

\title{Bandit-based Communication-Efficient Client Selection Strategies for Federated Learning
\thanks{This work was supported in part by NSF CCF grants \# 1850029 and \#2007834, a Carnegie Bosch Institute Research award, the CyLab IoT institute, the Korean Government doctoral study abroad fellowship (Yae Jee Cho) and the Barakat Presidential Fellowship (Samarth Gupta).}
}

\author{
\IEEEauthorblockN{Yae Jee Cho, Samarth Gupta, 
Gauri Joshi, 
Osman Ya\u{g}an\\
}\IEEEauthorblockA{Electrical \& Computer Engineering, Carnegie Mellon University, Pittsburgh, PA }}

\maketitle

\begin{abstract}
    Due to communication constraints and intermittent client availability in federated learning, only a subset of clients can participate in each training round. 
    While most prior works assume uniform and unbiased client selection, recent work on biased client selection~\cite{yjc2020csfl} has shown that selecting clients with higher local losses can improve error convergence speed. However, previously proposed biased selection strategies either require additional communication cost for evaluating the exact local loss or utilize stale local loss, which can even make the model diverge. In this paper, we present a bandit-based communication-efficient client selection strategy~\ucbc~that achieves faster convergence with lower communication overhead. We also demonstrate how client selection can be used to improve fairness. \\ 
\end{abstract}


\section{Introduction}

With increasing applications moving from data-center based training to edge-device training, federated learning (FL) \cite{mcmahan2017communication, kairouz2019advances} has been spotlighted as one of the powerful distributed optimization methods for on-device learning. FL enables a massive distributed network of devices (clients) to participate in the training process without data-sharing. In each communication round, a subset of selected clients perform local model training and send their updated models to a central aggregating server. 
Most previous works use unbiased client selection and model aggregation. However, due to the inherent data heterogeneity across clients, judicious use of bias in client selection presents an untapped opportunity to improve error convergence.

Recent works have \cite{yjc2020csfl, jack2019afl} shown that biasing client selection in FL towards clients with higher local loss achieves faster convergence compared to unbiased client selection. 
However the client selection strategies proposed in previous works either require the server to additionally communicate with clients to retrieve the accurate local loss values or use stale loss values received from selected clients from previous communication rounds. Communication is expensive in FL, and furthermore, we show that using stale loss values can lead to slower error convergence or even divergence. 

In this paper, we propose a bandit-based client selection strategy~\ucbc~that is communication-efficient and use the observed clients` local loss values more appropriately instead of using stale values. Moreover, we show that biased client selection can promote fairness, the uniformity of local loss performance across clients. To the best of our knowledge, there has been no work that proposes or evaluates biased client selection strategies in the context of fairness.

\section{Problem Formulation} \label{sec:pf}
\subsection{Federated Learning with Partial Device Participation}
The general FL framework FedAvg~\cite{mcmahan2017communication} aims to find the parameter vector $\wb$ that minimizes the following objective:
\begin{align}
\gf(\wb) &= \frac{1}{\sum_{k=1}^{K} \ldat}\sum_{\ic=1}^\ec \sum_{\xi \in \bdat} f(\wb, \xi) = \sum_{\ic=1}^\ec \rdat \lf(\wb) \label{eqn:global_objective}
\end{align}
with total $\ec$ clients, where client $\ic$ has a local dataset $\bdat$ consisting $|\bdat|=\ldat$ data samples. The term $f(\wb, \xi)$ is the composite loss function for sample $\xi$ and parameter vector $\wb$. The term $\rdat=\ldat/{\sum_{\ic=1}^{\ec}\ldat}$ is the fraction of data at the $k$-th client, and $\lf(\wb)=\frac{1}{|\bdat|}\sum_{\xi \in \bdat} f(\wb,\xi)$ is the local objective function of client $k$. \comment{In FL, the vectors $\wb^*$, and $\wb_{\ic}^*$ for $\ic = 1, \dots, \ec$ that minimize $\gf(\wb)$ and $\lf(\wb)$ respectively can be very different from each other, often resulting from large data heterogeneity across clients. We define $F^* = \min_{\wb} \gf(\wb) = F(\wb^*)$ and $F_k^* = \min_{\wb} \gf_k(\wb) = F_k(\wb_k^*)$.}

A central aggregating server optimizes the model parameter $\wb$ by 
selecting a subset of $\scc=C\ec$ clients for some fraction $0 < C < 1$ in each communication round (partial-device participation). 
Each selected client performs $\tau$ iterations of local SGD \cite{stich2018local,wang2018cooperative,yu2018parallel} and sends its locally updated model back to the server. Then, the server updates the global model using the local models and broadcasts the global model to a new set of active clients. Formally, we index the local SGD iterations with $t \geq 0$. The set of active clients at iteration $t$ is denoted by $\mathcal{S}^{(t)}$. Since active clients performs $\tau$ steps of local update, the active set $\mathcal{S}^{(t)}$ also remains constant for every $\tau$ iterations. That is, if $(t+1) \mod \tau = 0$, then $\mathcal{S}^{(t+1)}=\mathcal{S}^{(t+2)}=\dotsm=\mathcal{S}^{(t+\tau)}$. Accordingly, the update rule of FedAvg is as follows:
\begin{flalign}
\label{eqn:local_model_update}
&\wb_\ic^{\tip}= \nonumber \\  &\begin{cases}
\frac{1}{\scc} \sum_{j \in\st} \left(\wb_j^{\ti}- \eta_{t} g_j (\wb_j^{\ti},\xi_j^{(t)}) \right) \triangleq \owglb^{\tip}  ~~~\text{else}
\end{cases}
\end{flalign}
where $\wb_\ic^{\tip}$ is the local model of client $\ic$ at iteration $t$, $\lr$ is the learning rate, and $\sgrad(\wb_\ic^{\ti},\midat)=\frac{1}{b}\sum_{\xi \in \midat}\nabla \lff(\wb_\ic^{\ti}, \xi)$ is the stochastic gradient over mini-batch $\midat$ of size $b$ that is randomly sampled from client $\ic$'s local dataset $\mathcal{B}_k$. Moreover, $\owglb^{\tip}$ is the global model at the server.

\subsection{Biased Client Selection for Faster Convergence}
We define a client selection strategy $\pi$ that maps the parameter vector $\wb$ to a specific set of clients $\mathcal{S}(\pi,\wb)$. The baseline unbiased client selection strategy used in FedAvg~\cite{mcmahan2017communication} chooses clients in proportion to $\rdat$, denoted as $\pir$. It has been noted in \cite{yjc2020csfl, jack2019afl} that selecting clients with higher local loss at each communication round leads to faster convergence but incurs an error floor. To attain faster convergence than $\pir$, \cite{yjc2020csfl} proposes the power-of-d client selection scheme $\pipd$. Under the $\pipd$ scheme, the central server with $d > m$ clients obtains their local loss $\lf(\wb)$ for the current global model $\wb$. After doing so, the $\pipd$ selects the $m$ clients with largest local loss values in the next communication round. It has been seen in \cite{yjc2020csfl} that this scheme performs much better than the $\pir$ scheme for a variety of ML tasks. A drawback of $\pipd$ scheme is that it requires  additional communication, as the central server is required to poll $d$ clients before selecting clients for the next communication round.

To reduce this additional communication, it is desirable to have a proxy for the local loss $\lf(\wb)$ of each client $k$ available at the center. Motivated by this, \cite{yjc2020csfl}  proposes the $\piprr$ scheme, where the local loss $\lf(\wb)$ is approximated by the local loss of the client when it was last selected in the client selection procedure. However, this approximation can be misleading at times as the client loss evaluation can be noisy and stale. Due to these reasons, it was observed that in certain cases the $\piprr$ scheme does not have desirable convergence (See \Cref{fig:syn}). In this paper, we design a communication-efficient client selection strategy that enjoys faster convergence performance and robustness to the error floor compared to previously proposed client selection strategies.

\subsection{Fairness in Client Selection}
Fairness in FL has been studied only recently in the literature~\cite{li2019fair, mohri2019agnostic}, with the main goal of capturing the local accuracy discrepancies across clients for a trained global model. According to~\eqref{eqn:global_objective}, clients with larger $\rdat$ will intuitively yield lower local loss performance, and vice versa. However, if clients with small $\rdat$ perform significantly worse than the other clients, this can be \textit{unfair}~\cite{li2019fair, mohri2019agnostic}. Henceforth, we define \textit{client fairness} as the extent of identical local performance across clients for a single global model. We show that client fairness can be improved  by incorporating the estimated local loss values and client's selected frequency to the client selection scheme. Gaining perspective from wireless resource allocation~\cite{raj1998fair, car2012fair, lan@axiofairness}, we measure fairness by the Jain's index~\cite{raj1998fair} $J(\wb),~1/\ec\leq J(\wb)\leq1$ where $1$ is when all clients have the same performance. Fairness metric $J(\wb)$ is defined as:
\begin{align}
    J(\wb)=\frac{1}{\ec}\left[\sum_{\ic=1}^\ec\left(\frac{\lf(\wb)}{\sum_{i=1}^\ec F_i(\wb)}\right)^2\right]^{-1}
\end{align}
With this definition, we show that our proposed~\ucbc~gains both fairness and convergence speed compared to the unbiased and previously proposed biased client selection strategies.


\section{Client Selection with discounted UCB} \label{sec:cs}
In order to achieve faster convergence with low error floor, it is important to select clients with larger local loss (i.e., \emph{exploitation}) as that leads to faster convergence \cite{yjc2020csfl}. It is also important to ensure diversity (i.e., \emph{exploration}) in selection to achieve a lower error floor. Motivated by the fact that there is a exploration-exploitation trade-off, we propose the use of Multi-Armed Bandit (MAB) algorithms \cite{bubeck2012regret} for the problem of client-selection in FL. Since the local loss values of individual clients are non-stationary during training we make use of discounted MAB algorithms proposed in \cite{gari2008disucb}. We modify the discounted UCB algorithm \cite{gari2008disucb} to balance the exploration-exploitation trade-off in the client selection problem. We view the clients as the arms in the MAB problem and compute discounted cumulative local loss values of each client, $L_t(\gamma, k)$, and a discounted count of the number of times each client has been sampled, $N_t(\gamma,k)$, till communication round $t$.

\begin{figure*}[!t]
\centering
\begin{subfigure}{0.32\textwidth}
\centering
\includegraphics[width=1\textwidth]{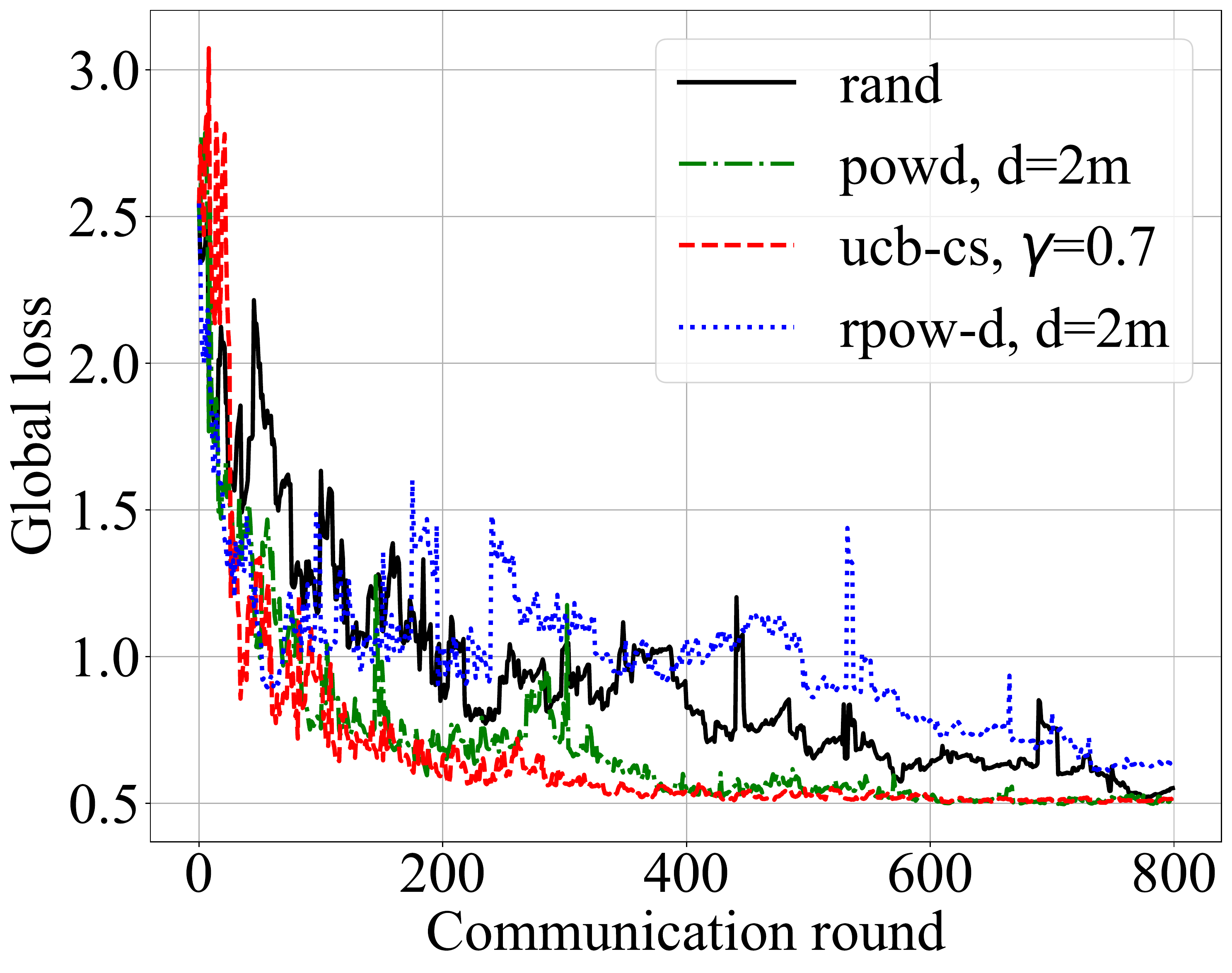} \caption{$m=1$}
\end{subfigure} \hspace{0.2em}
\begin{subfigure}{0.32\textwidth}
\centering
\includegraphics[width=1\textwidth]{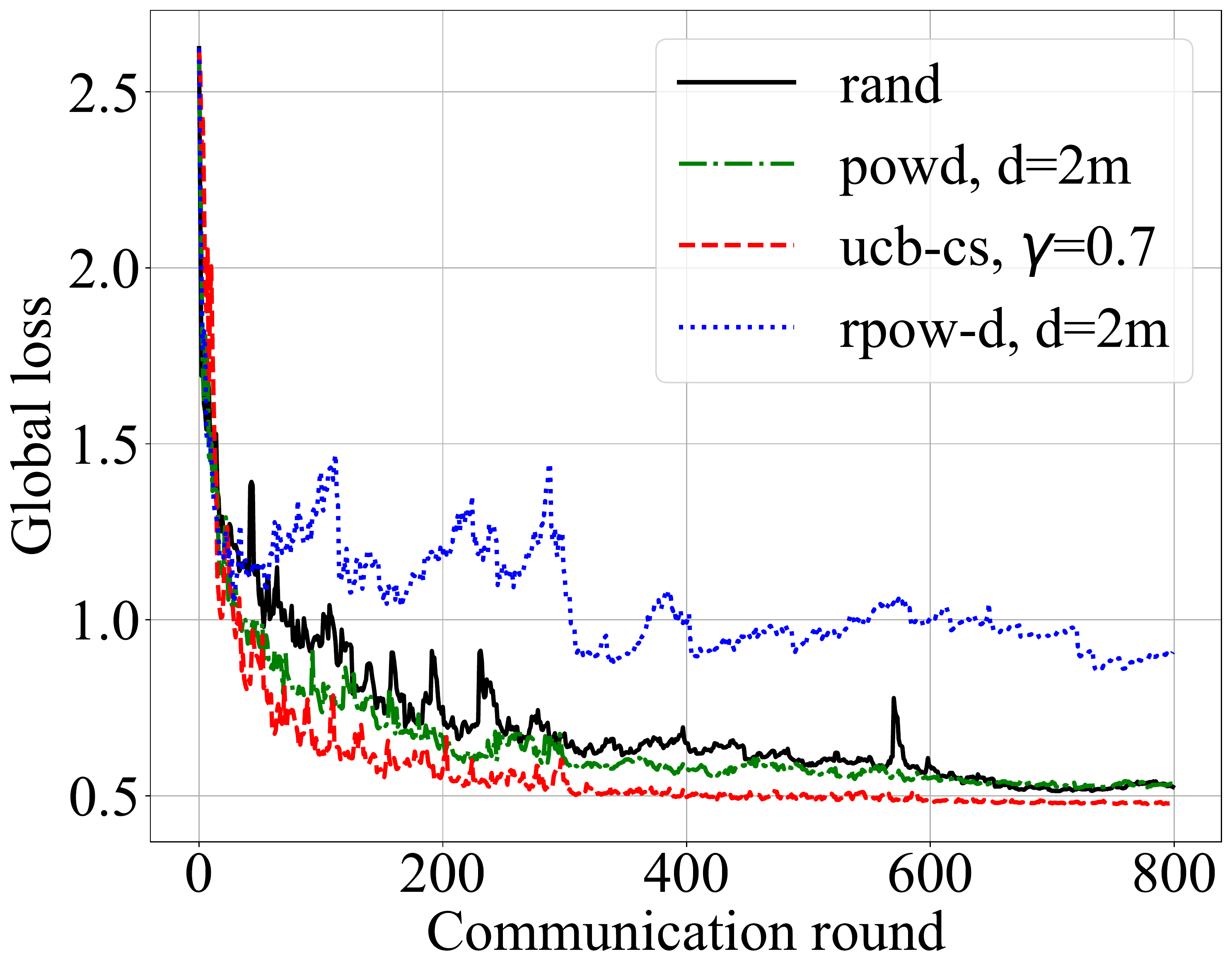}\caption{$m=2$}
\end{subfigure} \hspace{0.2em}
\begin{subfigure}{0.32\textwidth} 
\centering
\includegraphics[width=1\textwidth]{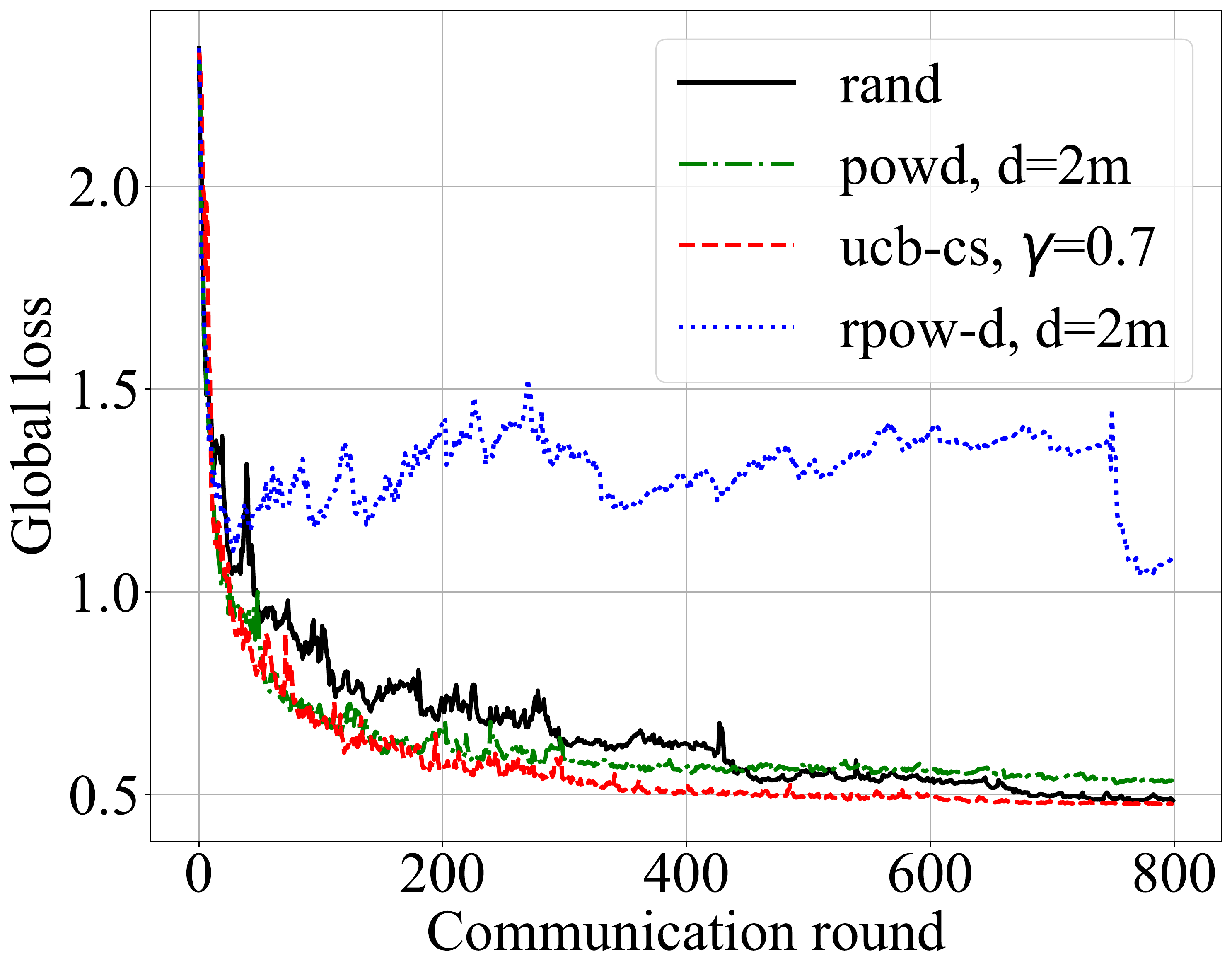}\caption{$m=3$}
\end{subfigure}
\caption{Global loss for logistic regression on the synthetic dataset, \texttt{Synthetic}(1,1), with $\pir$, $\pipd$, $\piprr$, and $\piucb$ for $d=2m,~\gamma=0.7$ where $\ec=30,~m\in\{1,2,3\}$.}
\label{fig:syn}
\vspace{-0.25cm}
\end{figure*}

Using these, we define the discounted UCB index for each client $k \in [K]$ at communication round $t$, and select the $m$ clients with largest discounted UCB indices. With $\mathcal{T}=\{\tau,~2\tau,~3\tau,~...,~\floor{t/\tau}\tau\}$ which are the set of time indices for the communication rounds until $t$, the discounted UCB indices $A_t(\gamma, k)$ are formally defined as 
\begin{align}
    A_t(\gamma,k)=\rdat(\underbrace{L_t(\gamma,k)/N_t(\gamma,k)}_{\text{exploitation}}+\underbrace{U_t(\gamma,k)}_{\text{exploration}}) \label{ucb}
\end{align}
where,
\begin{align}
    L_t(\gamma,k)&=\sum_{t'\in\mathcal{T}}\gamma^{t-t'}\mathbbm{1}_{\{\ic\in\mathcal{S}^{(t'-1)}\}}\frac{1}{\tau}\sum_{l=t'-\tau+1}^{t'}\sum_{\xi \in \xi_\ic^{(l)}}\frac{\lff(\wb_\ic^{(l)}, \xi)}{b}\\
    N_t(\gamma,k)&=\sum_{t'\in\mathcal{T}}\gamma^{t-t'}\mathbbm{1}_{\{\ic\in\mathcal{S}^{(t'-1)}\}}\\
    U_t(\gamma,k)&=\sqrt{2\sigma_t^2\log{T_t(\gamma)}/ N_t(\gamma,k)},~T_t(\gamma)=\sum_{t'\in\mathcal{T}}\gamma^{t-t'}
\end{align}
Here, $0 \leq \gamma \leq 1$ is a hyper-parameter that impacts the importance given to stale values. When $\gamma = 1$, all past local loss samples contribute equally in the calculation of $L_t(\gamma, k)$ and when $\gamma = 0$ only the latest local loss is used to estimate $L_t(\gamma, k)$. For $0 < \gamma < 1$, less weight is put upon stale values of local loss for the calculation of $L_t(\gamma,k)$. By doing so, we compute the estimate for local loss of client in a robust way to escape the noise in the latest evaluation and discount the stale values computed in the past. The exploration term $U_t(\gamma,k)$ gets larger by a factor of $\sigma_t$ especially for clients that have not been selected recently regardless of their local loss values. This forces $\ucbc$ to explore other clients that may have relatively smaller exploitation value. This not only can pull the algorithm away from generating the error floor by just selecting clients with estimated larger local loss, but also promotes fairness in terms of the number of times the client is explored. The parameter $\sigma_t$ is the maximum standard deviation in the local loss computed over the latest update of clients. Note that we multiply the dataset size ratio with the discounted UCB index, as we also want to sample clients proportional to their datasize for fast convergence and lower error floor. Details of~\ucbc~are presented in Algorithm~\ref{algo1}.

\begin{algorithm}
\caption{Pseudo code for \ucbc}\label{algo1}
\renewcommand{\algorithmicloop}{\textbf{Global server do}}
\begin{algorithmic}[1]
\State \textbf{Input}: $\scc,~\gamma$, $\rdat$ for $k\in[K]$
\State \textbf{Output}: $\st$ 
\State \textbf{Initialize}: empty sets $\st$ and list $\mathcal{A}$ of length $\ec$
\Loop
\State Receive $\frac{1}{\tau b}\sum_{l=t-\tau+1}^{t}\sum_{\xi \in \xi_\ic^{(l)}}\lff(\wb_\ic^{(l)}, \xi)$ from clients $\ic\in \mathcal{S}^{(t-1)}$ and calculate $A_t(\gamma,\ic)$ as (\ref{ucb})
\State Update $\mathcal{A}[k]=A_t(\gamma,\ic)$
\State Get $\st$ = \{$m$ clients with largest values in $\mathcal{A}$ (break ties randomly)\}
\State Discount elements in $\mathcal{A}$ by $\mathcal{A}=\gamma\mathcal{A}$
\EndLoop
\Return $\mathcal{S}^{\ti}$
\end{algorithmic}
\end{algorithm}

\section{Experiment Results} \label{sec:er}
We evaluate the proposed~\ucbc~with logistic regression on a heterogeneous synthetic federated dataset, \texttt{Synthetic}(1,1) \cite{sahu2019federated}, and DNN trained on a non-iid partitioned FMNIST dataset \cite{xiao2017fmnist}. For logistic regression, we assume in $K=30$ where the local dataset sizes follow the power law distribution. We set $b=50,~\tau=30$, and $\eta=0.05$, where $\eta$ is decayed to $\eta/2$ every 300 and 600 rounds. For DNN, we train a deep multi-layer perceptron network with two hidden layers on the FMNIST dataset. We construct the heterogeneous data partition amongst clients using the Dirichlet distribution $\text{Dir}_{K}(\alpha)$ \cite{hsu2019noniid}, where $\alpha$ determines the degree of the data heterogeneity across clients. Smaller $\alpha$ indicates larger data heterogeneity. For all experiments we use $b=64,~\tau=100$, and $\eta=0.005$, where $\eta$ is decayed by half for round 150. All experiments are conducted with clusters equipped with one NVIDIA TitanX GPU. The machines communicate amongst each other through Ethernet. The algorithms are implemented by PyTorch. For all results, the hyper-parameters $d$ and $\gamma$ are tuned for the best performance via grid search.

The training loss performance for the synthetic dataset simulation is presented in Fig.~\ref{fig:syn}. The~\ucbc~algorithm, $\piucb$, converges even faster than $\pipd$ without any error floor, and performs significantly better than $\pir$ and $\piprr$. The $\piprr$ selection policy performs worse than $\pir$, showing that using stale local losses for biased client selection can make the performance worse than the unbiased selection strategies. Additionally, in Table~\ref{tab:fair}, we show that the biased client selection strategies achieve notable higher fairness than the random selection strategy. While $\pipd$ is able to achieve higher fairness than $\piucb$, $\piucb$ shows a significant improvement in fairness even with low communication cost and robustness to the error floor in the training curve. Hence we show that $\piucb$ is efficient in the three important factors in FL: loss performance, fairness, and communication-efficiency.

\begin{figure}[!h] \centering
\includegraphics[width=0.46\textwidth]{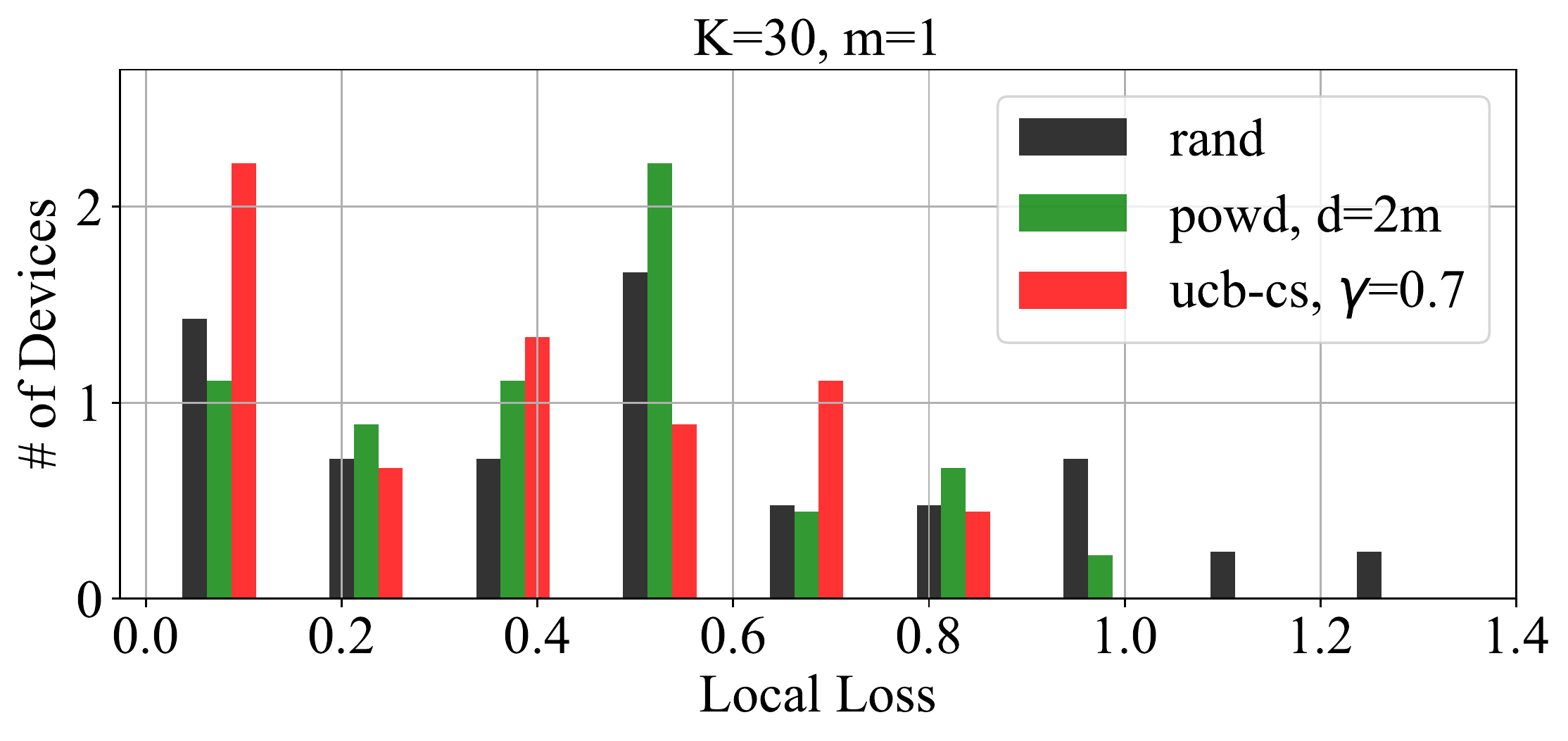} 
\caption{Histogram for the client`s individual loss performance after end of training for simulations presented in Fig.~\ref{fig:syn}(a).}
\label{fig:hist}
\vspace{-0.25cm}
\end{figure}

To dive deeper into the difference between $\pipd$ and $\piucb$, in Fig,~\ref{fig:hist} we present the local loss distribution across the clients at the end of training for the simulation in Fig.~\ref{fig:syn}(a). We show that both $\piucb$ and $\pipd$ is able to improve the worst performing client's local loss performance for $\pir$. While $\pipd$ is able to keep most of the clients in the approximately average range of performance of the local loss, $\piucb$ allows most of the clients to perform with the lowest local loss, skewing the local loss distribution across clients towards lower loss values. Hence from Fig.~\ref{fig:hist} we can see that $\pipd$ is valuing fairness over performance, whereas $\piucb$ is valuing performance slightly over fairness.

\begin{figure}[!t]
\centering
\begin{subfigure}{0.491\textwidth}
\centering
\includegraphics[width=1\textwidth]{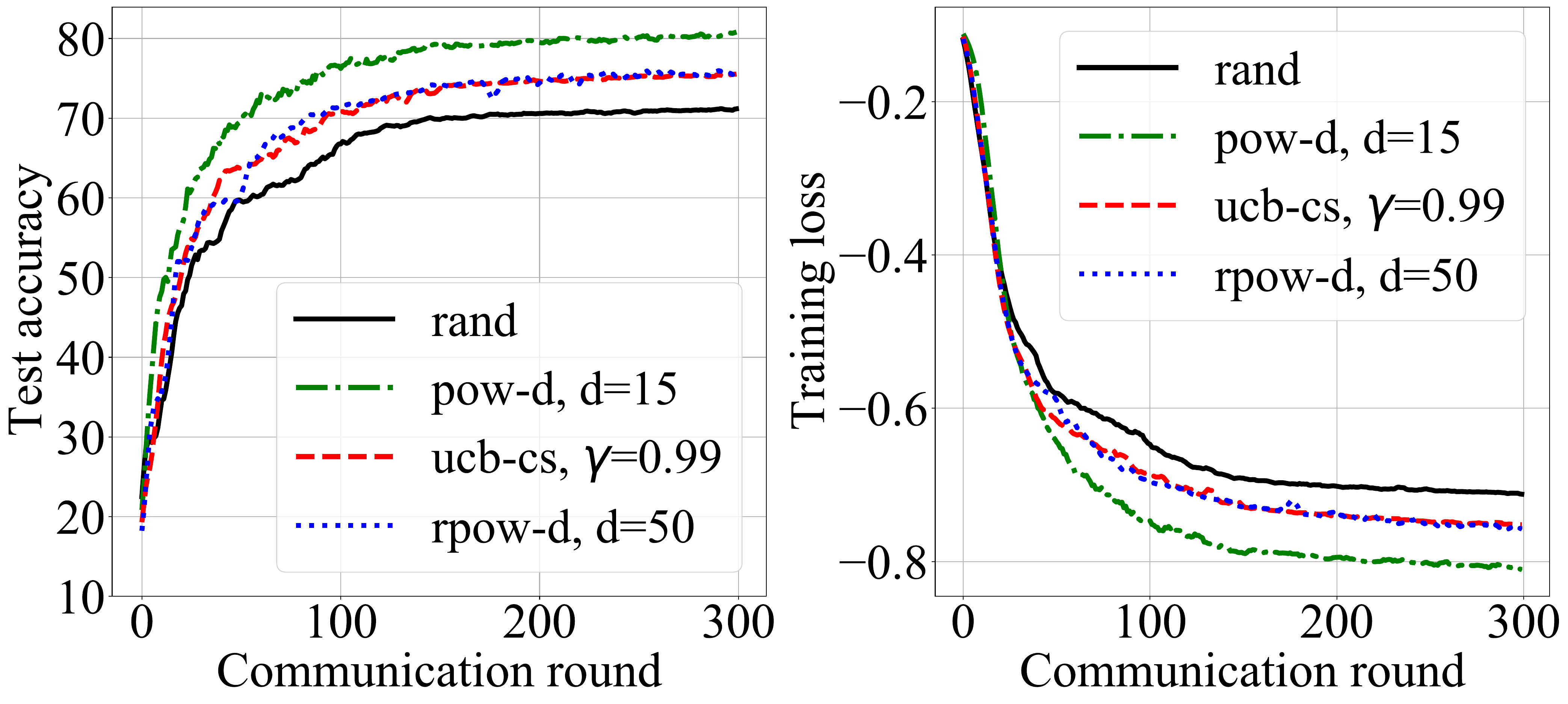} \caption{$\alpha=2$}
\end{subfigure}
\begin{subfigure}{0.491\textwidth}
\centering
\includegraphics[width=1\textwidth]{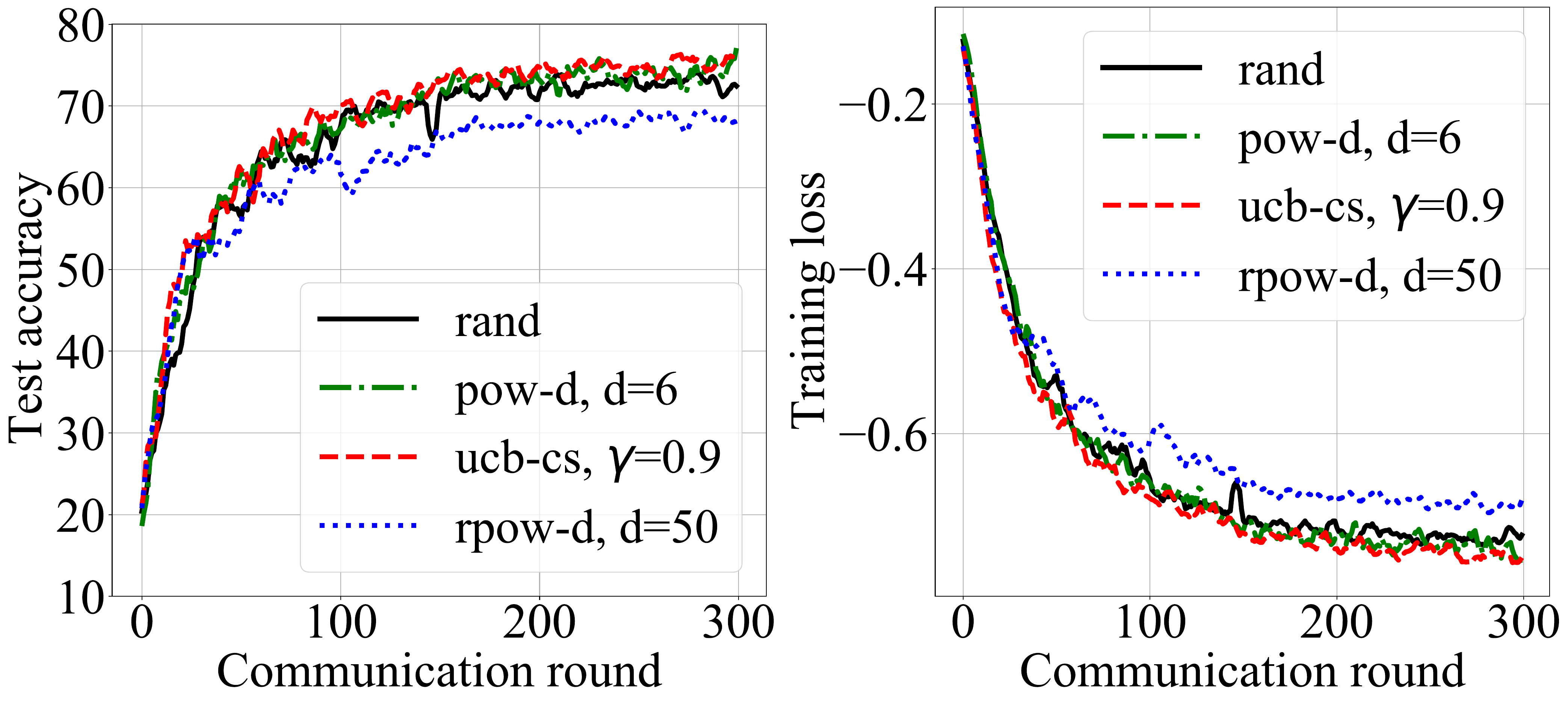}\caption{$\alpha=0.3$}
\end{subfigure} 
\caption{Test accuracy and training loss for $\pir,~\pipd$, $\piprr$, and $\piucb$ for $\ec=100,~C=0.03$ on the FMNIST dataset with mini-batch size $b=64$ and $\tau=100$.}
\label{fig:ml}
\vspace{-0.25cm}
\end{figure}

In Fig.~\ref{fig:ml}, the test accuracy and training loss for image classification on the FMNIST dataset via DNN are presented. For less data heterogeneity ($\alpha=2$), both $\piprr$ and $\piucb$ perform similarly with higher test accuracy and lower training loss than $\pir$. However, for larger data heterogeneity ($\alpha=0.3$), $\piprr$ performs worse than $\piucb$, showing that with large $\tau$ the estimated local loss values that $\piprr$ use becomes very stale,  worsening the performance in the presence of large data heterogeneity. On the other hand, $\piucb$ and $\pipd$ have similar empirical performance, which shows that $\piucb$'s use of discounted and accumulated local losses give an accurate representation of the client's actual local loss value.

\begin{table}[!t] \centering 
\small
\caption{Fairness values $J(\owglb^{(T)})$, where $T$ is the last communication round, for the scenarios in Fig.~\ref{fig:syn}.}
\begin{tabular}{@{}lccc@{}}\toprule
& $m=1$ & $m=2$ & $m=3$ \\\toprule
$\pir$ & $0.43$ & $0.29$ & $0.66$ \\  \hline
$\pipd$ & $0.75$ & $0.89$ & $0.91$ \\ \hline
$\piucb$ & $0.61$ & $0.61$ & $0.65$ \\ \hline 
$\piprr$ & $0.32$ & $0.52$ & $0.39$ \\
\bottomrule \label{tab:fair} \vspace{-2em}
\end{tabular}
\end{table}

\section{Concluding Remarks} \label{sec:cc}
In this paper, we propose a bandit-based communication-efficient client selection strategy,~\ucbc. It tackles the problem of communication-efficiency, noisy-stale estimates of local loss values, and error floor prevalent in biased client selection strategies from previous literature~\cite{yjc2020csfl, jack2019afl}. We discover that~\ucbc, with no additional communication compared to $\pir$, is robust to the error floor and gains convergence speed while mitigating the problem of staleness of observed local loss values. Moreover we show that~\ucbc~increases fairness, the uniformity of performance across different clients, compared to other known communication-efficient biased client selection strategies. Throughout this work, we assume that local losses corresponding to a model $\wb$ for each clients are independent of each other. However, in reality, similar clients might have similar local losses. We aim to use correlated multi-armed bandit algorithms \cite{gupta2020multiarmed, gupta2018correlated} in such setting to further improve the performance of \ucbc~for future work.
\bibliographystyle{IEEEtran}
\bibliography{bibfile.bib}

\end{document}